\def\year{2021}\relax
\documentclass[letterpaper]{article} 
\usepackage{aaai21}  
\usepackage{times}  
\usepackage{helvet} 
\usepackage{courier}  
\usepackage[hyphens]{url}  
\usepackage{graphicx} 
\usepackage{natbib}  
\usepackage{caption} 
\usepackage{xspace}

\newif\ifjournal

\journaltrue

\title{Strategy Proof Mechanisms for Facility Location with Capacity Limits}
\author {
\ifjournal
Toby Walsh
\else
Anonymous
\fi
\\
}

\affiliations{
\ifjournal
UNSW Sydney and Data61\\
tw@cse.unsw.edu.au
\else
    Affiliation \\
    Address \\
    anonymous@email.com
\fi
}

\newtheorem{mytheorem}{Theorem}
\newtheorem{myproposition}{Proposition}

\newtheorem{myexample}{Example}

\newcommand{\myproof}{\noindent {\bf Proof:\ \ }}
\newcommand{\mymax}{\mbox{\rm max}}

\newcommand{\myqed}{\mbox{$\diamond$}}
\newcommand{\myOmit}[1]{}

\newcommand{\innerpoint}{\mbox{\sc InnerPoint}\xspace}
\newcommand{\extendedendpoint}{\mbox{\sc ExtendedEndPoint}\xspace}
\newcommand{\percentile}{\mbox{\sc Percentile}\xspace}
\newcommand{\myendpoint}{\mbox{\sc EndPoint}\xspace}
\newcommand{\median}{\mbox{\sc Median}\xspace}
\newcommand{\leftmost}{\mbox{\sc Leftmost}\xspace}
\newcommand{\jleftkright}{\mbox{\sc $j$Left$k$Right}\xspace}

\newcommand{\twoleftpeaks}{\mbox{\sc TwoLeftPeaks}\xspace}
\newcommand{\tworightpeaks}{\mbox{\sc TwoRightPeaks}\xspace}

\newcommand{\threeleftpeaks}{\mbox{\sc ThreeLeftPeaks}\xspace}
\newcommand{\threerightpeaks}{\mbox{\sc ThreeRightPeaks}\xspace}
\newcommand{\firstsecond}{\mbox{\sc CapSD}\xspace}
\newcommand{\firstsecondthird}{\mbox{\sc CapaSD}\xspace}

\begin{document}

\maketitle

\begin{abstract}
An important feature of many real world facility location
problems are capacity limits on the facilities. We show here
how capacity constraints make it harder to design strategy proof
mechanisms for facility location, but counter-intuitively can improve
the guarantees on how well we can approximate the optimal solution. 
As a baseline, we begin by surveying what
is known about strategy proof mechanisms without capacity
limits. For example, we identify those strategy proof mechanisms
for locating one or two uncapacitated facilities
with bounded (in fact optimal) approximation ratios for the total and maximum
distance. In addition, we prove that, for two or more uncapacitated
facilities,  the only deterministic, anonymous and strategy proof mechanism
with a bounded approximation ratio
of the optimal maximum cost solution is the \myendpoint\ mechanism for
two facilities. We then consider the impact of adding capacity constraints to facilities. 
With two identical capacitated facilities and no spare capacity, we
prove a strong characterization result: 
a mechanism is anonymous, Pareto optimal and strategy proof
iff it is the \innerpoint\ mechanism. Spare capacity and facilities of
different size interfere with our ability to achieve Pareto optimality and
strategy proofness. In particular, with two facilities of different capacity, or two identical facilities
with spare capacity,
we prove that 
no mechanism is anonymous, Pareto optimal and strategy proof. 
By contrast, if we drop capacity limits completely, there are multiple anonymous, Pareto
optimal and strategy proof mechanisms for locating two uncapacitated
facilities. Additional facilities also interfere with our ability to
achieve Pareto optimal and strategy proofness. 
With three or more capacitated facilities, even if the facilities
are identical and there is no spare capacity, we prove that 
no mechanism is anonymous, Pareto optimal and strategy proof. 
Again by contrast, if we drop capacity limits, there are multiple 
anonymous, Pareto optimal and strategy proof mechanisms for
locating three facilities. 
Finally, for the all settings described so far in which no mechanisms exist,
we argue that there are multiple mechanisms which are anonymous and
strategy proof but not Pareto optimal, or Pareto optimal and
strategy proof but not anonymous, or anonymous and Pareto optimal
but not strategy proof. Hence, anonymity, Pareto optimality
and strategy proofness are a minimal combination of incompatible
axioms in all of these settings. 
\end{abstract}

\section{Introduction}

The facility location problem is a classic problem in social choice
that has attracted attention from researchers in several fields
including AI, Operations Research, and Game Theory
(e.g. \cite{flp, proportional,ft2010,egktps2011,ptacmtec2013,sfaaai15,fsaaai2015,gnpijcai18,mnwflp}).
Our goal here is to design mechanisms to locate
one or more facilities fairly and optimally
to serve a set of agents. The locations of the agents are privately
known to the agents themselves. We therefore wish to identify
strategy proof mechanisms that are likely to elicit the true locations of the agents and
minimize the total or maximum distance
of the agents from the facilities serving them. 
Facility location problems model many real world problems including 
locating schools, hospitals, warehouses,
and libraries. In many of these problems, facilities have capacity
limits \cite{capacitated2001,aclpwine2019,acllwaaai2020}.
A school has only places for a certain number of students,
a warehouse can only serve a given number of shops, 
a hospital has only a limited number of beds, etc.
Our contribution is to demonstrate that such
capacity constraints can make it harder to design strategy proof
mechanisms, but paradoxically can increase the welfare of the solutions
that we are guaranteed to return. 

As in much previous work on mechanism design for facility location
(e.g. \cite{ptacmtec2013}), we consider the one-dimensional setting.
This models a number of real world problems
such as locating wastewater plants along
a river, or warehouses along a highway. There are
also various non-geographical settings that can be viewed
as one-dimensional facility location problems (e.g. choosing
the temperature for a classroom, or selecting a committee to
represent people with different political views). In addition,
there are settings where we can use the one-dimensional
problem to solve more complex problems (e.g. decomposing the 
2-d rectilinear problem into a pair of 1-d problems).
Finally, the one-dimensional problem is the starting point to
consider more complex metrics (e.g., trees and networks) and provides
bounds on solutions for these more complex
settings (e.g. lower bounds for the 1-d problem provide lower bounds
for the 2-d problem).

\section{Formal Background}

We have $n$ agents located on the real line, and
wish to locate $m$ facilities on the real line to serve all the
agents. Each agent $i$ is at location $x_i$. We suppose agents
are ordered so that $x_1 \leq \ldots \leq x_n$. In the uncapacitated setting, 
an agent is served by the
nearest facility, and a solution is a location $y_j$ for each
facility $j$. 
In the capacitated setting, the $j$th facility can serve up to $c_j$
agents. We assume that $\sum_{j=1}^m c_j \geq n$ so that every agent
can be served. One special setting we consider is when there is no
spare capacity (i.e. $\sum_{j=1}^m c_j = n$). Another special setting
we consider is when facilities are identical (i.e. $c_i=c_j$ for all
$i<j)$. 
A solution in the capacitated setting is both a location $y_j$ for each
facility $j$, and an assignment of agents to facilities such that
the capacity limit $c_j$ for each facility is not exceeded. 
Let $a_i \in  [1,m]$ denote the facility serving agent $i$, and 
$N_j$ denote the set of agents assigned to facility $j$, i.e.,
$N_j = \{i | a_i = j\}$. Then the capacity constraints in the
capacitated setting ensure $|N_j| \leq c_j$ for all $j \in [1,m]$. 
We consider an utilitarian objective of the total distance,
$\sum_{i=1}^n | x_i - y_{a_i}|$ and an egalitarian
objective of the maximum distance, $\mymax_{i=1}^n | x_i -
y_{a_i}|$. Our goal is to minimize one of these two welfare objectives. 

We consider a number of mechanisms for the uncapacitated facility location
problem. With parameters
$p_1$ to $p_m$ with $0 \leq p_1 \leq \ldots \leq p_m \leq 1$,
the \percentile\ mechanism 
locates facility $j$ at $x_{1+\lfloor p_j (n-1) \rfloor}$ for $j  \in [1,m]$. 
For example, the \leftmost\
mechanism has $m=1$ and $p_1=0$, while the \median\ mechanism
has $m=1$ and $p_1=\frac{1}{2}$. 
The \myendpoint\ mechanism which locates facilities at the left and
rightmost agents has $m=2$, $p_1=0$ and $p_2=1$.
The \jleftkright\ mechanism locates 
$j$ facilities at the leftmost
$j$ distinct locations of the agents,
and $k$ facilities at the rightmost
$k$ distinct locations. If agents declare less than 
$j+k$ distinct locations, the mechanism puts multiple
facilities at a location. The \myendpoint\ mechanism is the
\jleftkright\ mechanism with $j=k=1$,
the \twoleftpeaks\ mechanism has $j=2$ and $k=0$,
the \tworightpeaks\ mechanism has $j=0$ and $k=2$,
the \threeleftpeaks\ mechanism has $j=3$ and $k=0$,
and the \threerightpeaks\ mechanism has $j=0$ and $k=3$. 

We also consider a number of mechanisms for the {\em capacitated}
problem. For two capacitated facilities, the \innerpoint\ mechanism locates 
one facility at $x_{c_1}$ serving the leftmost $c_1$ agents,
and the other facility at $x_{1+c_1}$ serving the remaining agents. 
The \extendedendpoint\ mechanism
is an extension of the \myendpoint\ mechanism to deal
with capacity limits that retains strategy proofness. It was first 
proposed in \cite{acllwaaai2020}. 
It can deal with facilities of different capacity, as 
well as with spare capacity. 
\ifjournal
It is defined formally as follows.

Let $f_{1}$ and $f_{2}$ be the output locations of the two facilities.
Define $X_{1}=\{i | x_{i}-x_{1}\leq \frac{1}{2}(x_{n}-x_{1})\}$ and $X_{2}=\{i | x_{n}-x_{i}<\frac{1}{2}(x_{n}-x_{1})\}$.
{If $|X_{1}|\geq |X_{2}|$, execute one of the following three cases.}
\begin{description}
\item[Case 1.] If $ |X_{1}| \leq c_{1}$ and $|X_{2}|\leq c_{2}$, $f_{1}=x_{1}$ and $f_{2}=x_{n}$. 
Agents in $X_{1}$ are allocated to $f_{1}$ and the others are
allocated to $f_{2}$ as in the \myendpoint\ mechanism.
\item[Case 2.] If $ |X_{1}| > c_{1}$ and $|X_{2}| \leq c_{2}$, $f_{1}=2x_{c_{1}+1}-x_{n}$ and $f_{2}=x_{n}$. Agents $\{1,\cdots, c_{1}\}$ are allocated to $f_{1}$ and the others are allocated to $f_{2}$.
\item[Case 3.] If $ |X_{1}| \leq c_{1}$ and $|X_{2}|> c_{2}$, $f_{1}=x_{1}$ and $f_{2}=2x_{n-c_{2}}-x_{1}$. Agents in $\{1,\cdots,n-c_{2}\}$ are allocated to $f_{1}$ and the others are allocated to $f_{2}$.
\end{description}
{If $|X_{1}| < |X_{2}|$, switch the roles of the two facilities in above cases and execute one of them.}

\fi
We extend the \percentile\ and \jleftkright\ mechanisms
to the capacitated setting by allocating agents left to right
to facilities ordered left to right up by their capacity. 

We consider three desirable properties of mechanisms for facility
location problems:
anonymity, Pareto optimality and strategy proofness.
Anonymity is a fundamental fairness property that requires 
all agents to be treated alike. 
Pareto optimality is one of the most fundamental normative properties
in economics. It demands that we cannot improve the solution so
one agent is better off without other agents being worse off. 
Finally, strategy proofness is a fundamental game theoretic property
that ensures agents have no incentive to act strategically and try to
manipulate the mechanisms. 

More fomally, a mechanism is {\em anonymous} iff permuting the agents does not
change the solution. 
For instance, the \median\ mechanism is anonymous. 
A mechanism is {\em Pareto optimal} iff 
it returns solutions that are always Pareto optimal. 
A solution is {\em Pareto optimal} iff there is no
other solution in which one agent travels a strictly shorter distance to the
facility serving them, and all other agents travel no greater
distance. For example,  the \median\ mechanism is
Pareto optimal. 
A mechanism is {\em strategy proof} 
iff no agent can mis-report and reduce their distance to travel to
the facility serving them. 
For example,  the \median\ mechanism is
strategy proof. 

Finally, we wil consider strategy proof
mechanisms that may not return the 
optimal solution. In particular, we will consider how well a mechanism approximates the
optimal possible welfare. A mechanism achieves an approximation ratio $\rho$ of the total (maximum)
distance iff the total (maximum) distance in any solution it returns
is at most $\rho$ times the optimal. In this case, we say that the mechanism
$\rho$-approximates the optimal total (maximum) distance. 
For example, the \leftmost\ mechanism 2-approximates the 
optimal maximum distance (Theorem 2.1 in \cite{ptacmtec2013}).

\section{Uncapacitated Facilities}

We begin by summarizing some existing results about strategy proof mechanisms
for the uncapacitated facility location problem, and by filling in a few
small gaps in the literature. This summary of uncapacitated results will serve as a baseline to
compare against when we add capacity constraints.

\subsection{Strategy Proofness and Anonymity}
 
With any number of facilities, every \percentile\ mechanisms is strategy proof 
(Theorem 1 in \cite{percentile}) and anonymous.  
Similarly, with any number of facilities $m$, 
any \jleftkright\ mechanism with $j+k=m$ 
is strategy proof and anonymous. 

\subsection{Pareto Optimality}

Pareto optimality is somewhat more difficult to achieve than
strategy proofness, especially as the
number of facilities increases. 
With a single facility, any \percentile\ mechanism
is Pareto optimal. 
With two facilities, the only
\percentile\ mechanism that is Pareto optimal is the \myendpoint\ 
mechanism. 
With three or more
facilities, no \percentile\ mechanism is Pareto optimal. 
On the other hand, with any number of facilities $m$ ($m \geq 1$), 
every \jleftkright\ mechanism with $j+k=m$ 
is Pareto optimal.

\subsection{Total Distance}

With a single facility, 
the \median\ mechanism returns the solution with optimal total
distance. With two facilities, no deterministic, anonymous and strategy proof mechanism
has a bounded approximation ratio for the total distance
except for the \myendpoint mechanism which has a bounded approximation ratio of 
$n-2$ (Theorem 3.1 in \cite{ft2013}).
With three or more facilities, no deterministic and strategy proof mechanism
at all has a bounded approximation ratio for the 
total distance  (Theorem 7.1 in \cite{ft2013}).

\subsection{Maximum Distance}

With a single facility, any \percentile\ mechanism
2-approximates the optimal maximum distance. 
Indeed no deterministic and strategy proof mechanism can have a better
approximation ratio (Theorem 3.2 of \cite{ptacmtec2013}). 
With two facilities, the \myendpoint\
mechanism also provides a 2-approximation of
the maximum distance. Again, no deterministic and strategy proof
mechanism can have a better approximation ratio (Corollary 4.4 of
\cite{ptacmtec2013}). 
In fact, the \myendpoint\ mechanism is the {\em only} \percentile\ mechanism for locating two uncapacitated
facilities with a bounded
approximation ratio for the maximum distance. 

\begin{mytheorem}
With two uncapacitated facilities, the only \percentile\
mechanism with a bounded approximation
ratio for the maximum distance is the \myendpoint\ mechanism. 
\end{mytheorem}
\myproof
There are three cases. In the first case, $p_1=p_2$.
Consider a problem with agents at 0 and 1. The optimal
maximum distance is 0 yet the \percentile\ mechanism returns an
allocation with maximum distance of 1.
In the second case, $0<p_1< p_2$. 
Consider a problem with $n=\lceil \frac{3}{p_2-p_1} \rceil$ agents.
Suppose 
one agent is at 0 and the rest are at 1. The optimal
maximum distance is 0 yet the \percentile\ mechanism returns an
allocation with both facilities at 1 giving
a maximum distance of 1.
In the third case, $0=p_1<p_2<1$. 
This case is symmetric to the second. 
\myqed

With three or more facilities, no \percentile\ mechanism 
has a bounded approximation ratio for
the maximum distance
irrespective of the parameters chosen. 

\begin{mytheorem}
With three or more uncapacitated facilities, no \percentile\
mechanism has a bounded approximation
ratio for the maximum distance.
\end{mytheorem}
\myproof
Consider three facilities.
There are five cases. In the first case, $p_1=p_2=p_3$. 
Consider a problem with agents at 0 and 1. The optimal
maximum distance is 0 yet the \percentile\ mechanism returns an
solution with maximum distance of 1.
In the second case, $p_1=p_2 < p_3$. 
Consider a problem with $n=\lceil \frac{3}{p_3-p_2} \rceil$ agents.
By definition, $n \geq 3$. Suppose 
$1+\lfloor p_2 (n-1) \rfloor$ agents are at 0,
one is at $\frac{1}{2}$ and the rest are at 1. The optimal
maximum distance is 0 yet the \percentile\ mechanism returns an
solution with two facilities at 0, and one at 1 giving
a maximum distance of $\frac{1}{2}$.
In the third case, $p_1< p_2 = p_3$. 
This case is symmetric to the second case.
In the fourth case, $p_1<p_2 < p_3$ and $p_3 - p_2 > p_2 - p_1$.
Consider a problem with $n=\lceil \frac{3}{p_2-p_1} \rceil$ agents.
Suppose 
$1+\lfloor p_1 (n-1) \rfloor$ agents are at 0,
one is at $\frac{1}{2}$ and the rest are at 1. The optimal
maximum distance is 0 yet the \percentile\ mechanism returns a
solution with one facility at 0, and two at 1 giving
a maximum distance of $\frac{1}{2}$.
In the fifth case, $p_1<p_2 < p_3$ and $p_3 - p_2 \leq p_2 - p_1$.
This case is symmetric to the fourth case.
The proof easily extends to four or more facilities.
\myqed

Actually, the last two results can be strengthened from 
any \percentile\ mechanism to any deterministic and strategy proof
mechanism and to fewer agents. We need the following simple proposition. 

\begin{myproposition} \label{totalmax}
A class of mechanisms has a bounded approximation ratio for 
the maximum distance iff the same class has a bounded approximation ratio
for the total distance. 
\end{myproposition}
\myproof
We exploit the following relation between the number of agents ($n$), 
the total distance ($d_{total}$), and the maximum distance
($d_{max}$): $ d_{total} \leq n d_{max}$. 
Therefore if $d_{max}$ is bounded then so is $n d_{max}$, and by
transitivity,  $d_{total}$ is bounded. 
For the reverse, we exploit the simple relation: 
$d_{max} \leq d_{total}$.
Therefore if $d_{total}$ is bounded then so is $d_{max}$. 
Hence, the maximum distance is bounded iff the total distance is too.
\myqed

By mapping across the corresponding results about total
cost from \cite{ft2013} using this simple proposition, 
we provide the following
characterization result for strategy proof mechanisms
with a bounded approximation for the maximum cost. 

\begin{mytheorem}
With two or more uncapacitated facilities, the only
deterministic, anonymous and strategy proof mechanism with a bounded approximation
ratio for the maximum distance is the \myendpoint\ mechanism (in the
case of two uncapacitated facilities). 
\end{mytheorem}
\myproof
This follows immediately from 
Proposition \ref{totalmax}, 
Theorem 3.1 in \cite{ft2013} which identifies \myendpoint\ as the only 
deterministic, anonymous and strategy mechanism
with bounded approximation for the total cost with two uncapacitated
facilities and five or more agents, and Theorem 7.1 in \cite{ft2013}
which proves that no such mechanism exists with $m$ 
uncapacitated facilities ($m \geq 3$) and $m+1$ or more agents. 
\myqed

\ifjournal
To summarize, with a single uncapacitated facility, any parameter for the \percentile\
mechanism provides a 2-approximation of the optimal maximum distance
(and  no deterministic and strategy proof mechanism
can do better); 
with two facilities, only the parameters $0$ and $1$ 
provide a 2-approximation (and 
again no deterministic and strategy proof mechanism
can do better); 
with three or more facilities whatever the parameters, or
with two facilities and any parameters besides $0$ and $1$, 
the approximation ratio is unbounded (however, 
no deterministic, anonymous and strategy proof mechanism
can bound the approximation ratio in this setting). 
\fi
Having summarized and modestly extended the literature on strategy proof mechanisms
for locating uncapacitated facilities, we are now in a position to consider
the impact of adding capacity constraints to facility
location problems. 

\section{Two Capacitated Facilities}

Since we suppose facilities have enough capacity to serve all agents,
the first non-trivial capacitated case to consider is two facilities. 
Capacity constraints can make strategy proofness
harder to achieve even in this simplest of settings. 
For instance, the \myendpoint\ mechanism stops being strategy
proof when we add capacity limits. In fact, the only \percentile\ mechanism that remains
strategy proof with the addition of capacity constraints is the
\innerpoint\ mechanism 
(Theorem 9 of \cite{acllwaaai2020}).  

Supposing we have two identical facilities
with no spare capacity, we now prove that 
the \innerpoint\ mechanism is in fact the {\em only} mechanism
that is anonymous, Pareto optimal and strategy proof. 
We contrast this strong characterization result with the uncapacitated problem 
where there are {\em multiple} mechanisms for locating two
uncapacitated facilities that are anonymous, Pareto optimal and strategy proof
(e.g. the  \twoleftpeaks\ or \myendpoint\ mechanisms). 

\begin{mytheorem} \label{two}
With $2k$ agents and two facilities of capacity $k$, 
a mechanism is anonymous, Pareto optimal and strategy proof iff it is the
\innerpoint\ mechanism.
\end{mytheorem}
\myproof
It is easy to see that the \innerpoint\ mechanism is
anonymous, Pareto and strategy proof. Therefore
it remains to show that if a mechanism 
is anonymous, Pareto optimal and strategy proof then it is the
\innerpoint\ mechanism.
We actually prove a stronger statement:
if a mechanism 
is anonymous, Pareto optimal and strategy proof for
$2k$ agents and two facilities of capacity $k$, or for $2k+1$
agents and a facility of capacity
$k$ on the left and another facility of capacity
$k+1$ on the right then that mechanism must be
the \innerpoint\ mechanism locating one facility at the $k$th
agent serving the leftmost $k$ agents, and the other 
facility at the $k+1$th agent serving the remaining agents. 
The proof uses induction on $k$, the capacity of the facilities.
We suppose agents report locations on $[0,1]$ and scale the
problem to ensure this if necessary. 

\ifjournal 
In the base case, $k=1$. There are two subcases. In the first subcase, we 
have just two agents and two facilities of capacity 1. 
The unique Pareto optimal solution locates a facility at the location
of each agent. A mechanism that does this is a strategy proof and
anonymous. This is also the solution return by the \innerpoint\
mechanism. In the second subcase, we have three agents, a facility of capacity 1 on
the left and a facility of capacity 2 on the right. 
Suppose the three agents are at $x_1 \leq x_2 \leq x_3$. 
The Pareto optimal solution set puts the smaller facility at $x_1$ serving the
leftmost agent and the larger facility somewhere in the interval
$[x_2,x_3]$ serving the other two agents. Now move the agent at $x_3$ to $x_2$. 
The unique Pareto optimal solution puts the smaller facility at $x_1$ serving the
leftmost agent and the larger facility at $x_2$ serving the two agents
there. We next move
the rightmost agent from $x_2$ back towards its original position at 
$x_3$. Let $x$ be the distance of the rightmost agent from $x_2$ so
that $x$ varies from 0 to $x_3-x_2$, and $f(x)$ be the distance of the
rightmost agent from the facility serving them. 
It is not hard to show that since the mechanism is strategy proof,
$f(x)$ must be a continuous function of
$x$. Any discontinuity would give the rightmost agent an opportunity
to mis-report  their location strategically and travel less distance. 
The location of the rightmost facility therefore tracks
continuously to the right, staying within the interval $[x_2,x_2+x]$
to ensure Pareto optimality. 
There are four scenarios for where the mechanism locates 
the rightmost facility as we vary $x$: (a) the larger facility remains at $x_2$
as with the \innerpoint\ mechanism, (b) the larger facility remains at
$x_2+x$,
(c) the larger facility tracks 
$x_2+x$ until some $x'$ with $x_2+x'<x_3$ after which the location
of the larger facility remains static
or (d) the larger facility at some point tracks behind and is strictly
between $x_2$ and $x_2+x$. 
Note that the larger facility cannot track in front of $x_2+x$
as this would not be Pareto optimal. 
In case (b), consider $x_1=\frac{1}{5}$, $x_2=\frac{2}{5}$, $x_3=1$ and $x=\frac{3}{5}$. 
Then the middle agent at $x_2$ can profitably misreport their
location as $0$. The leftmost facility will then be located 
at 0  serving the middle agent. The distance the 
middle agent  travels to be served thereby decreases from $\frac{3}{5}$ to $\frac{2}{5}$
violating the assumption that the mechanism is
strategy proof. 
In case (c), suppose $x_1=\frac{x}{2}$, $x_2=x$ then the agent at 
$x_2$ can profitably misreport their location as
$\frac{x}{2}$, contradicting the assumption that the
mechanism is strategy proof. 
In case (d), the larger  facility
tracks strictly behind $x_2+x$. By continuity arguments,
we can identify two values, $x=a$ and $x=b$ with $a<b$
such that when the rightmost agent is at 
$x_2+b$, the rightmost facility is located at $x_2+a$,
and when the rightmost agent is at 
$x_2+a$, the rightmost facility is located at $x_2+c$ where $c<a$.
Then if agents are at $x_1$, $x_2$ and $x_2+a$, the rightmost
agent at $x_2+a$ can profitably misreport their location as $x_2+b$. 
This violates the assumption that the mechanism is
strategy proof. Therefore the only case that does not
lead to a contradiction is case (a). That is, the mechanism acts like
the \innerpoint\ mechanism. 

\fi
\ifjournal In the step case, we \fi
suppose that the only anonymous, Pareto optimal and strategy proof
mechanism for $2k$ agents and two facilities of capacity $k$, or
for $2k+1$ agents and a facility of capacity $k$ on the left and a
facility of 
capacity $k+1$ on the right is the \innerpoint\ mechanism. We 
need to prove two subcases: the first subcase of $2k+2$ agents 
and two facilities of capacity $k+1$, and the second subcase
of $2k+3$ agents and a facility of capacity $k+1$ on the left and a
facility of 
capacity $k+2$ on the right. 
We consider the first subcase. 
\ifjournal 
Suppose the $2k+2$ agents are at $x_1 \leq x_2 \leq \ldots \leq x_{2k} \leq
x_{2k+1} \leq x_{2k+2}$. 
\fi
We move the leftmost agent at $x_1$ to 0 and suppose it is fixed and
served by the leftmost facility. We now have
a facility location problem with $2k+1$ agents, and a facility of capacity $k$ on left
and $k+1$ on the right. By the induction hypothesis, the only
anonymous, Pareto optimal and strategy proof mechanism
is the \innerpoint\ mechanism that locates one facility at $x_{k+1}$
serving agents located in the interval $[x_2,x_{k+1}]$, and the other
facility at $x_{k+2}$ serving the remaining agents. We next move the
leftmost agent from 0 back to $x_1$. 
\ifjournal By similar continuity arguments used in the base case, 
\fi 
the leftmost facility must remain at $x_{k+1}$. Continuity
arguments also prevent the rightmost facility moving away from
$x_{k+2}$ or for agents switching facility when $x_{k+2} > x_{k+1}$
as such a switch changes the distances traveled. If $x_{k+2}=x_{k+1}$
we don't care which facility serves which agent as the two facilities
are co-located and switching is irrelevant. Hence, 
with the leftmost agent back at $x_1$, the solution is that 
returned by the \innerpoint\ mechanism. 

This leaves the final subcase of $2k+3$ agents 
and a facility of capacity $k+1$ on the left and a 
facility of 
capacity $k+2$ on the right. 
\ifjournal
Suppose the agents are at $x_1 \leq x_2 \leq \ldots \leq x_{2k} \leq
x_{2k+1} \leq x_{2k+2} \leq x_{2k+3}$. 
\fi
We move the rightmost agent at $x_{2k+3}$ to 1 and suppose it is fixed
and served by the rightmost facility. 
We now have
a facility location problem with $2k+2$ agents, and two facilities of capacity $k+1$. By
the previous case, the only
anonymous, Pareto optimal and strategy proof mechanism for such a setting
is the \innerpoint\ mechanism that locates one facility at $x_{k+1}$
serving agents located in the interval $[x_1,x_{k+1}]$, and the other
facility at $x_{k+2}$ serving the remaining agents. We next move the
rightmost agent from 1 back to $x_{2k+3}$. By similar continuity
arguments, the rightmost facility must remain at $x_{k+2}$. Continuity
arguments also prevent the leftmost facility moving away from
$x_{k+1}$ or for agents to switch facility when $x_{k+2} > x_{k+1}$
and such a switch changes the distances traveled. Hence, 
with the rightmost agent back at $x_{2k+3}$, the solution is that 
returned by the \innerpoint\ mechanism. 
\myqed

It follows immediately that the \innerpoint\ mechanism
is the unique \percentile\ mechanism that is Pareto optimal. 
We next consider dropping in turn one of anonymity, Pareto optimality and strategy
proofness. If we drop anonymity,
then there are {\em multiple} other mechanisms 
besides the \innerpoint\ mechanism
that are Pareto optimal and strategy proof but not anonymous. 
\ifjournal For example, the \firstsecond\ mechanism is
a modified serial dictator mechanism for locating 
capacitated facilities that is Pareto optimal
and strategy proof but not anonymous. \fi

\begin{myexample}[\firstsecond\ mechanism]
Consider the following capacitated version of a serial
dictatorship. We order the agents in some arbitrary way. We then
locate the first facility at the position of the first agent in
this order. If the next agent in the order is at the same location and there 
is capacity remaining in the first facility, the we allocate this next
agent to the first facility and repeat. 
Otherwise we locate the second facility at the position of the next agent in the 
order. To finish the solution, we must allocate any remaining agents to
one of the two facilities to ensure
Pareto optimality. We consider the remaining agents in order,
allocating each to the nearest facility with remaining capacity. 
If an agent is equidistant from both facilities and both facilities have
spare capacity, we skip allocating them to a facility till
a final phase. This prevents spare capacity in a facility being
taken up by an agent that doesn't care to which facility it is
allocated, when that spare capacity could be better used by a later 
agent. Finally, we are left with unallocated agents that
are equidistant from the two facilities. We can allocate these agents
in any arbitrary way respecting capacities of the facilities. 
The resulting mechanism 
is Pareto optimal and strategy proof
but not anonymous. The mechanism can be easily extended to allocate
three or more facilities, respecting capacity constraints yet ensuring
Pareto optimality.
\end{myexample}

If we drop Pareto optimality, there are {\em multiple}
mechanisms besides the \innerpoint\ mechanism
that are anonymous and strategy proof but not Pareto optimal (e.g. any \percentile\
mechanism with $p_1=p_2$). 
And if we drop strategy proofness, there are {\em multiple} mechanisms 
besides the \innerpoint\ mechanism
that are anonymous and Pareto optimal but not strategy proof (e.g. the
\myendpoint\ and 
\tworightpeaks\ mechanisms). 
Hence, anonymity, Pareto optimality and strategy
proofness are the minimal combination of axioms
characterizing the \innerpoint\ mechanism. 

\section{Three Capacitated Facilities}

If we increase the number of facilities, strategy proofness
and Pareto optimality become harder to achieve simultaneously. 
In fact, even in the restricted setting of just three capacitated facilities of equal
size but no spare capacity, 
no mechanism can be simultaneously anonymous, Pareto optimal and
strategy proof. 
We again contrast this with the uncapacitated problem 
where there are {\em multiple} mechanisms for locating three
uncapacitated facilities that are anonymous, Pareto optimal and strategy proof
(e.g. the \threeleftpeaks\ and \threerightpeaks\ mechanisms). 
The following strong impossibility theorem demonstrates that
capacity limits make it impossible to achieve anonymity, Pareto
optimality and strategy proofness with three or more facilities. 
 
\begin{mytheorem}
With $km$ agents and $m$ facilities of capacity $k$ ($m \geq 3$ and $k
\geq 2$), 
no mechanism is anonymous, Pareto optimal and strategy proof. 
\end{mytheorem}
\myproof
We demonstrate the absence of an anonymous, strategy  
proof and optimal mechanism for 6 agents and 3 facilities
of capacity 2. The argument easily generalizes to more agents
and to more facilities. 
Suppose there is such a mechanism acting on 6 agents:
two at position 0, two at 10, and two at 20.
There is an unique Pareto optimal solution,
with one facility at 0, another at 10 and a third at 20.
Consider moving one of the agents, which we call $a$ from
its position at 10.
Let $f(x)$ be the function representing the distance of the facility
serving $a$ from their reported location $x$. Thus 
$f(10)=0$. As the mechanism 
locating facilities is strategy proof, 
$f(x)$ needs to be a continuous function of $x$. 
Indeed, it
is possible to show that if $x$ changes by some 
$\delta$ then $f(x)$ cannot change by more than $2\delta$
or strategic manipulation of the mechanism would be possible. 
Now suppose we move agent $a$ from position 10 
to 11. Then the facility serving $a$ cannot be too far
from $a$. Indeed, to ensure Pareto optimality, it
must be the case that one facility remains 
at 0, the other at 20, and the third lies somewhere
in the interval $[10,11]$. 

Suppose now the agents at 0 are fixed, but the other four agents are 
allowed to vary. Then we 
effectively have a two facility, four agent mechanism 
acting on the agents at 10, 11 and two at 20 which
locates the two rightmost facilities, while the leftmost facility
remains at 0. By Theorem \ref{two},
the only anonymous, Pareto optimal and  strategy proof 
mechanism for locating two facilities
is the \innerpoint\ mechanism. This locates the 
two rightmost facilities at 11 and 20. 
Suppose now that the agents at 20 are fixed, 
but the other four agents are 
allowed to vary. Then we again 
effectively have a two facility, four agent mechanism acting on 
agents at 10, 11 and two at 0 which
locates the two leftmost facilities, while the rightmost
facility remains at 20. By Theorem \ref{two},
this must be the \innerpoint\ mechanism which locates the 
two leftmost facilities at 0 and 10. This is a contradiction as 
the middle facility cannot simultaneously 
by at position 10 and 11. Therefore our assumption that
there is an anonymous, Pareto optimal and strategy proof
mechanism for three capacitated facilities must be incorrect. 
\myqed

It follows immediately that no \percentile\ mechanism
for locating three or more capacitated facilities is Pareto optimal. 
\ifjournal
If we drop anonymity, there are 
{\em multiple} mechanisms 
that are Pareto optimal and strategy proof (e.g. any
\firstsecondthird\ mechanism). 
Similarly, if we drop Pareto optimality, there are {\em multiple} mechanisms 
that are anonymous and strategy proof (e.g. any \percentile\
mechanism with $p_1=p_2=p_3$). 
And if we drop strategy proofness, there are {\em multiple} mechanisms 
that are anonymous and Pareto optimal (e.g. 
\threeleftpeaks, and \threerightpeaks\ mechanisms). 
\fi
Hence, anonymity, Pareto optimality and strategy
proofness are a minimal combination of incompatible axioms
for locating three or more capacitated facilities. 

\section{Welfare Bounds}

We next consider the impact of capacity constraints
on the ability of strategy proof mechanisms to approximate the optimal solution. 
With no spare capacity, we can give a tight
lower bound on the approximation ratio for the 
total distance. No deterministic and strategy proof mechanism
can do better than this bound.

\begin{mytheorem}
For $2k$ agents, 
any deterministic and strategy proof mechanism
to locate two facilities of capacity $k$ has
an approximation ratio for the total distance of at least
$k-1$. 
\end{mytheorem}
\myproof
Suppose there exists a mechanism with a smaller approximation ratio
than $k-1$. 
A mechanism is partially group strategy proof iff
no group of agents at the same location
can individually benefit if they misreport simultaneously.
As in the uncapacitated setting (Lemma 2.4
in \cite{ft2013}), a simple transitivity 
proof demonstrates that in the two capacitated facility game, 
a strategy proof mechanism is also partially group strategy proof
\cite{proportional}. 
Since our mechanism is strategy proof by assumption, it is also
partially group strategy proof. 
And since it has a bounded approximation ratio,
it satisfies unanimity. 
Consider $2k$ agents at 0. By unanimity, 
the mechanism 
returns the solution with both facilities at 0. 
Consider $k+1$ agents simultaneously
moving from 0 to location $x$ for $0 \leq x \leq 1$. 
Since the mechanism is partial group strategy proof and unanimous,
one of three cases must occur. 
In the first case, both facilities remain at 0. 
In the second case, both facilities track $x$ to location 1.
In the third case, both facilities track $x$ to location
$a$ and then remain stationary. 
The total distance is lowest in the second case and is 
$k-1$. However, even this best case 
contradicts the 
assumption that there exists a mechanism with a smaller approximation ratio
than $k-1$ since the optimal solution has 
facilities at 0 and 1 with a total distance of just 1.
Note that this lower bound on the approximation
ratio 
\ifjournal would apply to any metric space, not just the line. 
\fi
\myqed

With two identical facilities of capacity $k$ and no spare capacity, the \innerpoint\ mechanism
has an approximation ratio for the total cost of at most $k-1$
(Theorem 5 in \cite{acllwaaai2020}).
Hence, the \innerpoint\ mechanism is optimal. 
No other deterministic and strategy proof mechanism
can have a better approximation ratio. 
We contrast this with the uncapacitated setting
where the lower bound for any deterministic and strategy proof
mechanism is $n-2$ which is nearly twice as big
(Corollary 3.2 in \cite{ft2013}). Counter-intuitively adding capacity
constraints improves the guarantee on the approximation 
of the optimal solution. 

We also prove a tight lower bound on the approximation
ratio for the maximum distance. 
No deterministic and strategy proof mechanism
can do better than this bound. Interestingly, 
this is the same lower bound as in the uncapacitated setting. 

\begin{mytheorem} \label{five}
With two or more facilities of capacity $k$, and no spare capacity,
any deterministic 
and strategy proof mechanism 
has an approximation ratio for the maximum distance of at least 2. 
\end{mytheorem}
\myproof
Suppose we have an approximation ratio that is at most $a$ with $a
\geq 1$. 
Consider two
agents at $4a$, and the other two
agents in $[0,1]$. 
The optimal allocation has a maximum distance of at most $\frac{1}{2}$. 
Hence, to meet the approximation ratio, the mechanism will 
return a solution with a maximum distance of $\frac{a}{2}$ or less.
Therefore, the two agents in $[0,1]$ will be allocated a facility
at or to the left of position $1+\frac{a}{2}$, and the two agents at $4a$ to a facility in $[\frac{7a}{2},\frac{9a}{2}]$. 
Now we can view the allocation of the two agents in $[0,1]$ as 
a problem of locating a single facility of unbounded capacity using
a deterministic and strategy proof mechanism with bounded
approximation ratio for the maximum distance. 
And by Theorem 3.2 of \cite{ptacmtec2013}, this leads to a
contradiction 
for $a<2$. Hence, the approximation ratio $a$ is at least 2. 
The proof extends to additional agents and facilities easily.
\myqed

As the \innerpoint\ mechanism achieves an approximation
ratio of at most 2 \cite{acllwaaai2020}, it is {optimal}. 
No deterministic and strategy proof mechanism is guaranteed
to return better quality solutions. It is perhaps surprising that,  with two capacitated
facilities, we
can approximate the maximum distance so well. 
For the more relaxed problem of approximating the
maximum distance with a single uncapacitated facility, the best possible
deterministic and strategy proof mechanism already has an approximation
ratio of 2. Despite complicating the problem with an additional
facility, and
with capacity limits, we can still do just as well. 

\section{Spare Capacity}

So far, we have supposed that there is no spare capacity
in the problem. We now relax this assumption. 
We observe that this can make Pareto optimality and strategy proofness
harder to achieve simultaneously even with just two facilities.

\begin{mytheorem}
No mechanism is anonymous, Pareto optimal and
strategy proof
when we are locating $m$ facilities of equal capacity $c$ and there is a spare capacity of 
size $k$ where $m \geq 2$, $c > 2$ and $m(c-2) > k \geq 1$.
\end{mytheorem}
\myproof
Consider $m=2$, $c=3$ and $k=1$. 
Suppose we have three agents at 0 and two at 1. 
Then the unique Pareto optimal solution puts one 
facility at 0 and the other at 1. 
Now suppose one agent at 0 is fixed. This
modified problem is equivalent to a 
problem with four agents and two facilities of capacity 2. 
The only anonymous, Pareto optimal and strategy proof and 
mechanism for such a four agent problem is the \innerpoint\ mechanism.
Suppose we move one of the agents at 0 to 1. 
The \innerpoint\ mechanism now 
locates both facilities at 1. This is not Pareto optimal. 
Our modified problem has two agents at 0 and three at 1. The unique Pareto optimal
solution of this modified problem has one facility at 0 and the other at 1. 
The construction easily extends to larger $c$, $m$ and $k$. 
For example, for larger $c$, we put additional agents at both 0 and 1
while for larger $m$, we put additional agents at 2. 
\myqed

Note that if the spare capacity is sufficiently large then there are
mechanisms that are anonymous, Pareto 
optimal and strategy proof. For example, when the spare capacity is $m(c-1)$, we have
only $m$ agents for the $m$ facilities. We can then simply locate a facility at each
agent. Another edge case is three agents and two facilities of 
capacity 2. Consider the mechanism
that locates one facility at the leftmost agent, the other at the rightmost
agent, and allocates the middle agent to whichever facility is nearest. 
This mechanism is anonymous, Pareto optimal and strategy proof

\ifjournal
Note also that if we drop anonymity, there are 
{\em multiple} mechanisms 
that are Pareto optimal and strategy proof despite
the presecne of spare capacity (e.g. any \firstsecond\ mechanism). 
Similarly, if we drop Pareto optimality, there are {\em multiple} mechanisms 
that are anonymous and strategy proof despite the presence of spare capacity (e.g. any \percentile\
mechanism with $p_1=p_2=p_3$). 
And if we drop strategy proofness, there are {\em multiple} mechanisms 
that are anonymous and Pareto optimal despite the presence of spare capacity (e.g. 
\twoleftpeaks, and \tworightpeaks\ mechanisms). 
\fi
Hence, anonymity, Pareto optimality and strategy
proofness are a minimal combination of incompatible axioms
in the presence of spare capacity. 

Counter-intuitively, when we introduce 
spare capacity, the lower bound on approximating the optimal total distance 
increases. 
More capacity means we may have a less good approximation of the optimal solution. 
More precisely, increasing the spare capacity increases
the lower bound on the optimal total 
distance linearly from $\frac{n}{2}-1$ to
$n-2$. Eventually and unsurprisingly
it matches the lower bound of $n-2$ seen in the uncapacitated
setting (Corollary 3.2 in  \cite{ft2013}).
Note that we cannot directly apply lower bounds from 
the uncapacitated setting to our capacitated
problem as the space of 
mechanisms in the capacitated setting is imposing and thus different
to that in the uncapacitated setting \cite{ft2010}. 
In the capacitated
setting, we force an agent
to be served by a particular facility (an ``imposing'' mechanism)
while in the uncapacitated
setting agents can visit whichever facility is nearest their
true location. 

\begin{mytheorem} \label{seven}
Any deterministic and strategy proof mechanism
for $n$ agents ($n > 2$) to locate two facilities both of capacity $n-k$ or larger
for $\frac{n}{2} \geq k \geq 1$
has an approximation ratio for the total distance of at least $n-k-1$. 
\end{mytheorem}
\myproof
Consider $k=1$. 
Suppose there exists a deterministic and strategy proof
mechanism
with an approximation ratio of less than $n-2$. 
As the approximation ratio is bounded, the mechanism
satisfies a generalized form of unanimity
in which any zero distance solution is returned. 
Consider $n-1$ agents at $0$, 
and one agent at $-n^2$. 
The optimal total distance is 0 so any mechanism
with an approximation ratio of less than $n-2$
puts one facility at $0$ and the other at $-n^2$, both
facilities serving the agents at their immediate location. 
Suppose we now move one agent at 0 rightwards
to location $x$ for $x \geq 0$. 
We then have one agent at $-n^2$, $n-2$ agents at 0 and
one agent at $x$. 
To ensure strategy proofness and unanimity, one of three cases occurs.
In the first, the facility originally at 0 tracks $x$ and continues to
serve this agent. In the second, the facility at 0 stays at 0 and
continues to serve the agent at $x$. 
In the third, the facility at 0 tracks $x$ until some location $a>0$,
and then remains fixed. 
In the second case, the lower bound
on the approximation ratio is violated when $x=n^2$. 
In the third case, if we increase $x$ to $a+n^2$ with 
the facility remaining at $a$, 
the approximation ratio is again surely violated. 
Hence, we 
can only be in the first case, with the facility
tracking $x$.
Suppose $x=1$.
To meet the lower bound, the agents at 0 must continue
to be served by the same facility.
By similar arguments, starting from 
$n-1$ agents at 1 and one agent at $-n^2$ and 
moving $n-2$ agents at 1 back to 0, 
we can conclude that the $n-2$ agents
at 0 must be served by the facility at 1. 
But the optimal solution puts one facility at $-n^2$ and the other at 0. The
solution constructed by our mechanism with one facility at 1 violates
the lower bound wherever the second facility is located. 
Hence we have a contradiction that a deterministic and strategy proof
mechanism exists which beats the lower bound. 
For larger $k$, we construct a similar argument
around the possible location of facilities with
$k$ agents at $-n^2$, 
$n-k-1$ agents at $0$, 
and one agent at $x$ for $x \geq 0$. 
Note that for $n$ even and $k=\frac{n}{2}$, the lower 
bound derived here of $n-k-1$ equals the 
lower bounds 
of $\frac{n}{2}-1$ derived in Theorem 1. 
\myqed

\ifjournal
For two facilities of capacity $n-1$ or greater,
the \myendpoint\ mechanism (which locates a facility at the leftmost and
rightmost locations serving whichever agents are nearest)
achieves this lower bound and so is optimal. No other
deterministic and strategy proof mechanism
can have a better approximation ratio for the total distance. 
We see therefore that increasing the spare capacity turns the problem
inside out -- the optimal mechanism goes from
the \innerpoint\ mechanism that locates facilities on the inside around the
median agents when there is no spare capacity to 
the \myendpoint\ mechanism that locates facilities at the end points of the reported
locations when there is a large amount of spare capacity. 
\fi
It follows immediately that the \extendedendpoint\ mechanism, 
which has an approximation ratio for the
total distance of  $\frac{3n}{2}$ (Theorem 11 in
\cite{acllwaaai2020}), is asymptotically optimal. 
It is an interesting open question to close the gap
between $n-k-1$ and $\frac{3n}{2}$. 

For the maximum distance, we can easily adjust the proof of Theorem
\ref{five} to show that, even with spare capacity, the approximation
ratio for the maximum distance is again at least 2. 
It immediately follows that the \extendedendpoint\ mechanism, 
which achieves an approximation ratio of $4$ (Theorem 12 in
\cite{acllwaaai2020}),  is asymptotically optimal. 
It is also an interesting open question to close the gap
between $2$ and $4$. 

\section{Unequal Capacity}

So far, we have supposed that all facilities have the same capacity. 
What happens if facilities can have different capacity to each other?
We will show that unsurprisingly this can make it harder to achieve
Pareto optimality and strategy proofness simultaneously. For instance, with two facilities of different
capacities,  there is no anonymous, Pareto optimal and strategy proof
mechanism. 

\begin{mytheorem}
With two (or more) facilities of two
(or more) different capacities, no mechanism is anonymous, Pareto optimal and strategy proof,
with or without space capacity, provided the capacity of the smallest facility is at least 2. 
\end{mytheorem}
\myproof
Consider no spare capacity, one facility of capacity 3 and the other
of capacity 2. Suppose we have three agents at 0 and two at 1. 
Then the unique Pareto optimal solution puts the
facility of capacity 3 at 0 and the facility of capacity 2 at 1. 
Now suppose we fix one agent at 0. This modified
problem is equivalent to a 
problem with four agents and two facilities of capacity 2. 
The only anonymous, Pareto optimal and strategy proof
mechanism for such a four agent problem is the \innerpoint\ mechanism.
Suppose we move one of the agents at 0 to 1. 
The \innerpoint\ mechanism locates both facilities at 1. This is not Pareto optimal. 
The unique Pareto optimal solution of our modified problem has the 
facility of capacity 2 at 0 and the facility of capacity 3 at 1. 
The construction easily extends to larger capacities, to more than
two facilities, and to problems with spare capacity. 
\myqed

\ifjournal
An edge case is when we have two facilities and one has just capacity
for a single agent.  More generally, suppose we have $m+1$ agents, one facility of 
capacity $m$ and the other of capacity 1 where $m>1$.
Consider the mechanism that locates
the largest facility at the leftmost agent, 
and the other at the rightmost agent when
the $m$th agent from the left is nearer the leftmost agent,
and otherwise locates the largest facility at the rightmost agent
and the other facility at the leftmost agent. Agents are allocated
to the nearest facility. 
This mechanism is anonymous, strategy proof and Pareto optimal. 
\fi

\ifjournal
Note also that if we drop anonymity, there are 
{\em multiple} mechanisms 
that are Pareto optimal and strategy proof despite
the presence of facilities of unequal capacity (e.g. any \firstsecond\ mechanism). 
Similarly, if we drop Pareto optimality, there are {\em multiple} mechanisms 
that are anonymous and strategy proof despite the presence of facilities of unequal
capacity (e.g. any \percentile\
mechanism with $p_1=p_2=p_3$). 
And if we drop strategy proofness, there are {\em multiple} mechanisms 
that are anonymous and Pareto optimal despite the presecne of facilities of unequal 
capacity (e.g. 
\twoleftpeaks, and \tworightpeaks\ mechanisms). 
\fi
Hence, anonymity, Pareto optimality and strategy
proofness are a minimal combination of incompatible axioms
when facilities can have different capacity. 

As might be expected, the more equal the capacity of the different
facilities, the better the guarantee on the approximability of the
optimal solution. 

\begin{mytheorem}
Any deterministic and strategy proof mechanism
for $n$ agents ($n > 2$) to locate two facilities of capacity $k$ and $n-k$ 
with $\frac{n}{2} \geq k \geq 1$
has an approximation ratio for the total cost of at least $n-k-1$. 
\end{mytheorem}
\myproof
The argument is essentially the same as in the proof of Theorem \ref{seven}
\myqed

It again immediately follows from this result that the \extendedendpoint\ mechanism, 
which achieves an approximation ratio of $\frac{3n}{2}$ (Theorem 11 in
\cite{acllwaaai2020}) with facilities of different capacities, is asymptotically optimal. 
It is an interesting open problem to close
the gap between the upper and lower bounds. 

For the maximum distance, we can easily adjust the proof of Theorem
\ref{five} to show that, even with facilities of different capacities, the approximation
ratio for the maximum distance is at least 2. 
It follows that the \extendedendpoint\ mechanism, 
which achieves an approximation ratio of $4$ (Theorem 12 in
\cite{acllwaaai2020}),  is asymptotically optimal. 
It is again an interesting open problem to close
the gap between the upper and lower bounds.

\section{Conclusions}

We have studied the impact of capacity constraints on
mechanisms for facility location considering three important
axiomatic properties: anonymity which is a fundamental fairness
property,
Pareto optimality which is a fundamental efficiency property,
and strategy proofness which is a fundamental property about
incentives to report sincerely. As a baseline, we began by surveying what
is known about strategy proof mechanisms without capacity
limits. For example, we identified  those mechanisms
for locating one or two uncapacitated facilities
with bounded (in fact optimal) approximation ratios for the total and maximum
distance. In addition, we filled a small gap in the literature by
proving that, with three or more uncapacitated facilities,
the only deterministic, anonymous and strategy proof mechanism with a
bounded approximation ratio for the maximum distance is the
\myendpoint\ mechanism. 

Our main contribution, however, is in the design of strategy proof
mechanism for locating facilites with capacity limits on the number
of agents that can be served. 
We provided a comprehensive characterization of the anonymous,
Pareto optimal and strategy proof mechanisms for locating capacitated
facilities. We also identified tight bounds on how well such
mechanisms can approximate the optimal total or maximum distance. 
In general, and as might be expected, capacity constraints make it more 
difficult to design strategy proof mechanisms for facility
locations. However, a little 
unexpectedly, capacity constraints can result in better guarantees on the
quality of the approximate solutions returned. 
\ifjournal

There are many directions for future work. For example, randomization is a useful tool to enable
mechanisms to ensure desirable axiomatic properties like
anonymity and strategy proofness. It would therefore be interesting
to extend our analysis from purely deterministic to randomized 
mechanisms for capacitated facility location.  
As a second example,
many problems are not one dimensional. Can we extend these results 
to trees, networks, and
two-dimensional rectilinear or Euclidean metrics?
As a third example, 
there is a dual class of obnoxious facility location
problems where agents wish to be as far as possible
from the facility such as a rubbish dump, nuclear power station, or
prison. 
There are also mixed facility location problems where some agents
wish to be close to the facility and others far away
such as a playground or cell phone tower. 
It would be interesting to consider the impact of capacity
constraints on such problems. 

\fi

\eject

\bibliographystyle{aaai21}
\bibliography{/Users/tw/Documents/biblio/a-z,/Users/tw/Documents/biblio/a-z2,/Users/tw/Documents/biblio/pub,/Users/tw/Documents/biblio/pub2}

\end{document}